\documentclass[sigconf]{acmart}

\AtBeginDocument{%
  \providecommand\BibTeX{{%
    \normalfont B\kern-0.5em{\scshape i\kern-0.25em b}\kern-0.8em\TeX}}}

\copyrightyear{2021} 
\acmYear{2021} 
\setcopyright{acmcopyright}\acmConference[KDD '21]{Proceedings of the 27th ACM SIGKDD Conference on Knowledge Discovery and Data Mining}{August 14--18, 2021}{Virtual Event, Singapore}
\acmBooktitle{Proceedings of the 27th ACM SIGKDD Conference on Knowledge Discovery and Data Mining (KDD '21), August 14--18, 2021, Virtual Event, Singapore}
\acmPrice{15.00}
\acmDOI{10.1145/3447548.3467307}
\acmISBN{978-1-4503-8332-5/21/08}

\usepackage{booktabs}       
\usepackage{amsfonts}       
\usepackage{nicefrac}       

\usepackage{graphicx}
\usepackage{amsmath}
\usepackage{color}
\usepackage{mathrsfs}
\usepackage{amsthm}
\usepackage{multirow}
\usepackage{color,soul}
\usepackage{subfig}

\newtheorem{theorem}{Fact}

\DeclareMathOperator*{\argmin}{argmin}

\newcommand{\ie}{i.e.}
\newcommand{\eg}{e.g.}

\usepackage{array}
\newcolumntype{x}[1]{>{\centering\arraybackslash\hspace{0pt}}p{#1}}

\newcommand\blfootnote[1]{%
  \begingroup
  \renewcommand\thefootnote{}\footnote{#1}%
  \addtocounter{footnote}{-1}%
  \endgroup
}

\newcommand{\paragraphskip}{0.55em}
\newcommand{\captionskip}{-0.7em}

\begin{document}

\title{Why Attentions May Not Be Interpretable?}



\author{Bing Bai$^{1*}$, Jian Liang$^{2*}$,  Guanhua Zhang$^1$, Hao Li$^1$, Kun Bai$^1$, Fei Wang$^3$}
\affiliation{%
  \institution{$^1$Tencent Inc., China\quad$^2$Alibaba Group, China\\
  $^3$Department of Population Health Sciences, Weill Cornell Medicine, USA}
}
\email{{icebai, guanhzhang, leehaoli, kunbai}@tencent.com}
\email{xuelang.lj@alibaba-inc.com, few2001@med.cornell.edu}

\renewcommand{\authors}{Bing Bai, Jian Liang, Guanhua Zhang, Hao Li, Kun Bai, and Fei Wang}
\renewcommand{\shortauthors}{Bing Bai, Jian Liang, et al.}

\begin{abstract}
Attention-based methods have played important roles in model interpretations, where the calculated attention weights are expected to highlight the critical parts of inputs~(\eg, keywords in sentences). However, recent research found that attention-as-importance interpretations often do not work as we expected. For example, learned attention weights sometimes highlight less meaningful tokens like ``\texttt{[SEP]}'', ``\texttt{,}'', and ``\texttt{.}'', and are frequently uncorrelated with other feature importance indicators like gradient-based measures. A recent debate over whether attention is an explanation or not has drawn considerable interest.
In this paper, we demonstrate that one root cause of this phenomenon is the \emph{combinatorial shortcuts}, which means that, in addition to the highlighted parts, the attention weights themselves may carry extra information that could be utilized by downstream models after attention layers. As a result, the attention weights are no longer pure importance indicators.
We theoretically analyze combinatorial shortcuts, design one intuitive experiment to show their existence, and propose two methods to mitigate this issue. We conduct empirical studies on attention-based interpretation models. The results show that the proposed methods can effectively improve the interpretability of attention mechanisms.\blfootnote{$*$ Equal contributions from both authors. Dr. Liang participated in this work when he was at Tencent Inc.}
\end{abstract}



\maketitle

\section{Introduction}
\label{sec:intro}

Interpretation for machine learning models has gained increasing interest and becomes necessary in practice as the industry rapidly embraces machine learning technologies. 
Model interpretation explains how models make decisions, which is particularly essential in mission-critical domains where the accountability and transparency of the decision-making process are crucial, such as medicine~\citep{wang2019should}, security~\citep{chakraborti2019explicability}, and criminal justice~\citep{lipton2018mythos}.

\begin{figure}[t!]
  \centering
  \includegraphics[width=\columnwidth]{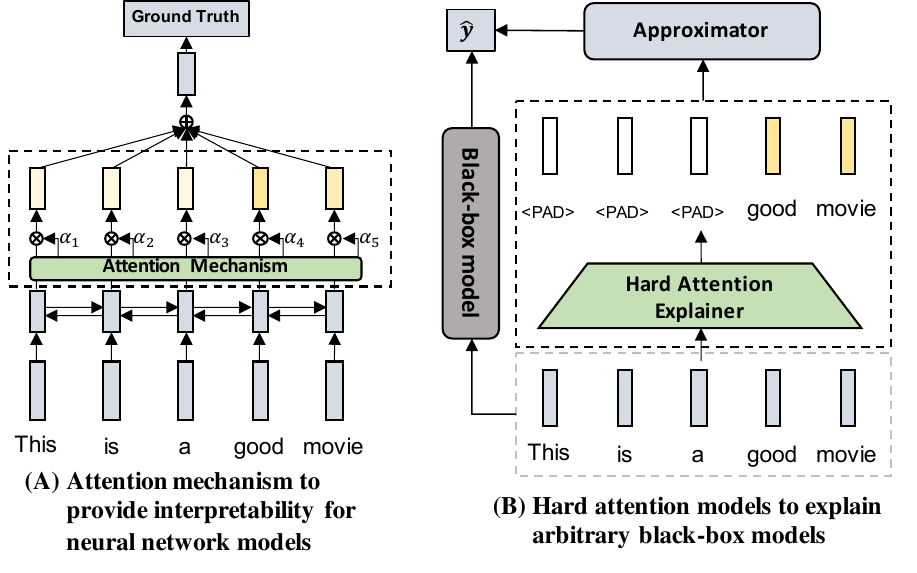}
  \vspace{\captionskip}
  \caption{The use of attention for model interpretations. With specific architectures, attention mechanisms can both provide interpretability for neural networks, and select features to explain black-box models.}
  \label{fig:illustration}
\end{figure}

Attention mechanisms have played important roles in model interpretations. 
As shown in Figure~\ref{fig:illustration}, attention mechanism as building blocks in neural networks are often applied to provide interpretability along with improving performance~\citep{choi2016retain,vaswani2017attention,wang2016attention},
and hard attention-based models are also used in explaining arbitrary black-box models' predictions, by selecting a fixed number of input components to approximate the predictions~\citep{chen2018learning,liang2020adversarial,belinkov2019analysis}.
More concretely, we formalize the attention mechanisms as $\text{Attention}(Q,K,V)=\text{Mask}(Q,K) \odot V$\footnote{Frequently a sum pooling operator is applied after the Hadamard product operator to obtain a single fixed-length representation. However, sometimes models other than simple pooling are applied~\citep{bang2019explaining,chen2018learning,zhu2017learning}. So we use the most general form here.} as in~\citep{vaswani2017attention}, where $Q$ is the query and $\langle K, V \rangle$ are the key-value pairs. 
$\text{Mask}(\cdot,\cdot)$ maps the query and keys to the attention weights (denoted as \emph{masks} in this paper), and then the \emph{masks} filter the information of $V$.
In other words, the \emph{masks} are expected to represent the importance of different parts of $V$~(\eg, words of a sentence, pixels of an image) and highlight those the models focus on to make decisions.


However, recent research suggested that the feature parts highlighted by attention mechanisms do not necessarily correlate with intuitively greater importance on the final predictions. 
For example, \citet{clark2019does} found that a surprisingly large amount of BERT’s attention focuses on less meaningful tokens like ``\texttt{[SEP]}'', ``\texttt{,}'', and ``\texttt{.}''.
Moreover, many researchers have provided evidence to support or refute the interpretability of the attention mechanisms. There have been lots of debates on whether attention is an explanation or not recently~\citep{jain2019attention,serrano2019attention,wiegreffe2019attention}. 

In this paper, we propose that a root cause hindering the interpretability of attention mechanisms is \textbf{\emph{combinatorial shortcuts}}. As mentioned earlier, we expect that the results of attention mechanisms mainly contain information from the highlighted parts of $V$, which is a critical assumption for attention-based interpretations' effectiveness. 
However, as the results are products of the masks and $V$, we find that the \emph{masks} themselves can carry extra information other than the highlighted parts of $V$, which the downstream parts of models could utilize.
As a result, the calculated masks may work as another kind of \emph{``encoding layers''} rather than providing pure feature importance.
For an extreme example, in a (binary) text classification task, the attention mechanisms could highlight the first word for positive cases and highlight the last one for negative cases, regardless of the word semantics. The downstream parts of attention layers can then predict the label by checking whether the first or the last word is highlighted, which may give good accuracy scores but completely fail to provide interpretability\footnote{One may argue that for the case where sum pooling is applied, we lose the positional information, and thus the intuitive case described above may not hold. 
However, since (1)~the distributions of different positions are not the same, (2)~positional encodings~\citep{vaswani2017attention} have also been used widely, the case is still possible.}.

We further study the effectiveness of attention-based interpretations and dive into the combinatorial shortcut problem. From the perspective of causal effect estimations, we first analyze the difference between coventional attention mechanisms and ideal interpretations theoretically, and then show the existence of combinatorial shortcuts through a constructive experiment. Based on the observations, we propose two methods to mitigate the issue, \ie, random attention pretraining and instance weighting for mask-neutral learning.
Without loss of generality, we examine the effectiveness of the proposed methods with an end-to-end attention-based model-interpretation approach, \ie, L2X~\citep{chen2018learning}, which can select a given number of input components to explain arbitrary black-box models.
Experimental results show that the proposed methods can successfully mitigate the adverse impact of combinatorial shortcuts and improve explanation performance.

\section{Related Work}
\label{sec:related_work}

\vspace{\paragraphskip} \noindent \textbf{Attention mechanisms for model interpretations} \ 
Attention mechanisms have been widely adopted in natural language processing~\citep{bahdanau2015neural,vinyals2015grammar}, computer vision~\citep{fu2016aligning,li2019attention}, recommendations~\citep{bai2020csrn,zhang2020general} and so on. 
Attention mechanisms have been used to explain how models make decisions by exhibiting the importance distribution over inputs~\citep{choi2016retain,martins2016softmax,wang2016attention}, which we can regard as a kind of model-specific interpretation.
Besides, there are also attention-based methods for model-agnostic interpretations. 
For example, L2X~\citep{chen2018learning} is a hard attention model~\citep{xu2015show} that employs Gumbel-softmax~\citep{jang2017categorical} for instancewise feature selection. 
VIBI~\citep{bang2019explaining} improved L2X to encourage the briefness of the learned explanation by adding a constraint for the feature scores to a global prior. \citet{liang2020adversarial} and \citet{yu2019rethinking} improved attention-style model interpretation methods through adversarial training to encourage the gap between the predictability of selected/unselected features.

However, there has been a debate on the interpretability of attention mechanisms recently.
\citet{jain2019attention} suggested that ``attention is not explanation'' by finding that the attention weights are frequently uncorrelated with other feature importance indicators like gradient-based measures.
On the other side, \citet{wiegreffe2019attention} argued that ``attention is not not-explanation'' by challenging many assumptions underlying \citet{jain2019attention} and suggested that they did not disprove the usefulness of attention mechanisms for explainability.
\citet{serrano2019attention} applied a different analysis based on intermediate representation erasure and found that while attention noisily predicts input components' overall importance to a model, it is by no means a fail-safe indicator.

In this work, we take another perspective on this problem called \emph{combinatorial shortcuts}, and show that it can provide one root cause of the phenomenon. We analyze why combinatorial shortcuts exist and propose theoretically guaranteed methods to alleviate the problem.

\vspace{\paragraphskip} \noindent \textbf{Causal effect estimations} \ 
Causal effect is an important concept for quantitative empirical analyses.
The causal effect of one treatment, $E$, over another, $C$, is defined as the difference between what would have happened if a particular unit had been exposed to $E$ and what would have happened if the unit had been exposed to $C$~\citep{rubin1974estimating}.
Randomized experiments, where the experimental units across the treatment groups are randomly allocated, play a critical role in causal inference. However, when randomized experiments are infeasible, researchers have to resort to nonrandomized data from surveys, censuses, and administrative records~\citep{winship1999estimation}, and there may be some other variables controlling the treatment allocation process in such data. For example, consider the causal inference between uranium mining and health. Ideally, the treatment (uranium mining) should be randomly allocated. However, mining workers are usually stronger people among all humans in the world. If some appropriate measures are absent, we may draw biased conclusions like ``uranium mining has no adverse health impact because the average life span of uranium mine workers is not shorter than that of ordinary people''. 
For recovering causal effects from nonrandomized data, instance weighting-based approaches have been used widely~\citep{ertefaie2010comparing,rosenbaum1983central,winship1999estimation}.

\section{Combinatorial Shortcuts}
\label{sec:problem}

\begin{figure*}[ht]
  \centering
  \includegraphics[width=0.825\textwidth]{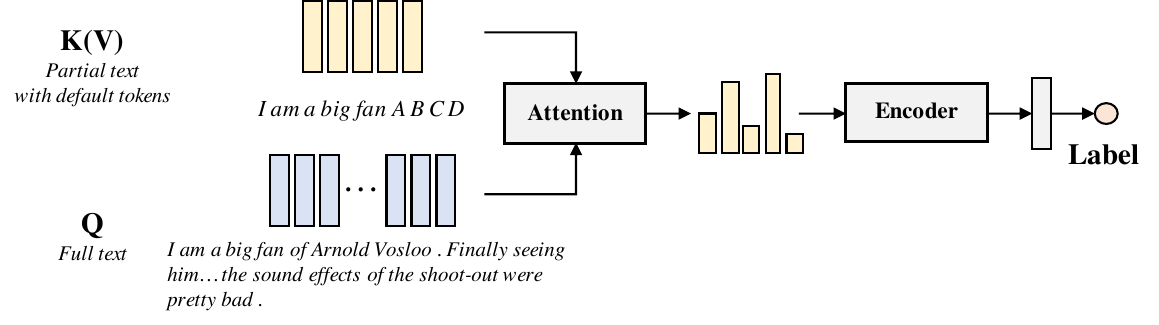}
  \vspace{\captionskip}
  \caption{The structure of the model used in this demo experiment. The attention is restricted to the first five tokens of sentences and four default tokens that appear in all samples. After the \emph{Attention} module, the weighted sequences are further encoded to a fixed-length vector with the \emph{Encoder} module.}
  \label{fig:demonstration}
\end{figure*}

\begin{table*}[ht]
  \centering
  \caption{Experiments about how much attention is put to default tokens out of all attention weights. Both Pooling and RCNN are tested as the \emph{Encoder} module in Figure~\ref{fig:demonstration}. Note that we report the results on the training set to demonstrate how models fit the data.}
  \label{tab:demonstration}
  \vspace{\captionskip}
  \begin{tabular}{cx{1.4cm}x{1.0cm}x{1.4cm}x{1.4cm}x{1.4cm}x{1.4cm}x{1.4cm}}
    \toprule
    \multirow{2}{*}{No.} & \multirow{2}{*}{Encoder} & \multirow{2}{*}{Label} &    \multicolumn{5}{c}{Attention to default tokens} \\
     & & & \texttt{A} & \texttt{B} & \texttt{C} & \texttt{D} & Total\\
    \midrule
    (1) & \multirow{2}{*}{Pooling} & \texttt{pos} & 68.0\% & 0.1\% & 0.0\% & 0.2\% & 68.3\% \\
    (2) & & \texttt{neg} & 0.1\% & 36.6\% & 38.3\% & 18.5\% & 93.5\% \\
    \midrule
    (3) & \multirow{2}{*}{RCNN} & \texttt{pos} & 16.1\% & 0.2\% & 0.1\% & 0.6\% & 17.0\% \\
    (4) & & \texttt{neg} & 1.3\% & 20.2\% & 50.4\% & 13.3\% & 85.2\% \\
    \bottomrule
  \end{tabular}
\end{table*}

In this section, by showing the difference between attention mechanisms and ideal explanations, we analyze why attention mechanisms become less interpretable from the perspective of causal effect estimations, and conduct an experiment to demonstrate the existence of combinatorial shortcuts.

\subsection{The difference between attention mechanisms and ideal explanations}
\label{sec:gap_theory}

Firstly, we analyze what ideal explanations are. Although the concept of ``explanation'' may differ in different situations, following the idea of ``finding the most dominating feature components'', we define the ideal explanations as follows in this paper.
We use uppercase letters to represent random variables, and lowercase letters to represent specific values of random variables.
Assume that we have samples drawn independently and identically distributed~(i.i.d.) from a distribution with domain $\mathcal{X} \times \mathcal{Y}$, where $\mathcal{X}$ is the feature domain, and $\mathcal{Y}$ is the label domain\footnote{Note that here we use $\mathcal{X}$ to represent the features for the convention. $X$ is the same as the value $V$ introduced in the Introduction. Besides, the labels could be either from the real world for explaining the real world, or from some specific models for explaining given black-box models.}.
Additionally, we assume that the mask is drawn from a distribution with domain $\mathcal{M}$. Usually, $\mathcal{M}$ is under some constraints for briefness, for example, only being able to select a fixed number of features or being non-negative and summing to $1$.
Given any sample $\langle x, y \rangle \sim \mathcal{X} \times \mathcal{Y}$, for $m_1 \sim \mathcal{M}$ and $m_2 \sim \mathcal{M}$, if $\mathcal{L}\big(\mathbb{E}(Y|x \odot m_1), y\big) < \mathcal{L}\big(\mathbb{E}(Y|x \odot m_2), y\big)$ where $\mathcal{L}(\cdot)$ is the loss function and $\mathbb{E}(\cdot)$ calculates the expectation, we say that for this sample, $m_1$ is superior to $m_2$ in term of interpretability. 
If an unbiased estimation of $\mathbb{E}(Y|X \odot M)$ is available, the best mask for sample $\langle x, y \rangle$ that can select the most informative features can be obtained by solving $\argmin_m \mathcal{L}\big(\mathbb{E}(Y|x \odot m), y\big)$.
As a conclusion, under the definitions above, the ideal explanation $m^*$ for sample $\langle x, y \rangle$ is $\argmin_m \mathcal{L}\big(\mathbb{E}(Y|x \odot m), y\big)$ with an unbiased estimation of $\mathbb{E}(Y|X \odot M)$.

In practice, we often need to train models to estimate $\mathbb{E}(Y|X \odot M)$. 
Ideally, if the data~(combinations of $X$ and $M$, as well as the label $Y$) is exhaustive and the model is consistent, we can train a model to obtain an unbiased estimation of $\mathbb{E}(Y|X \odot M)$ following the empirical risk minimization principle~\citep{fan2005improved,vapnik1992principles}.
Nevertheless, it is impossible to exhaust all combinations of $X$ and $M$. Taking a step back, from the perspective of causal effect estimations, we could consider different $M$ as different treatment, and randomized combinations for $X$ and $M$ can still be proven to give unbiased estimations on expectations~\citep{rubin1974estimating}.

However, attention mechanisms do not work in this way. Considering the downstream part of attention models, \ie, the part estimating the function $\mathbb{E}(Y|X \odot M)$, we can find that it receives highly selective combinations of $X$ and $M$. The used mask $M$ during the training procedure is a mapping from query and keys, making the used mask for samples highly related to the feature $X$ (and $Y$ as well).
Consequently, the training procedure of the attention mechanism produces a nonrandomized experiment~\citep{shadish2008can}. Therefore the model cannot learn unbiased $\mathbb{E}(Y|X \odot M)$. In turn, the attention mechanism will try to select biased features to adapt the biased estimations to minimize the overall loss functions, and thus fail to highlight the essential features.
As a result, the attention mechanism and downstream part may cooperate and find unexpected ways to fit the data, \eg, highlighting the first word for positive cases and the last word for negative cases, ultimately failing to provide interpretability.
This paper denotes the effects of nonrandomized combination for $X$ and $M$ that hinders the interpretability of attention mechanisms as \emph{combinatorial shortcuts}.

\subsection{Experimental demonstration for the combinatorial shortcuts}
\label{sec:demo_design}

To intuitively demonstrate the existence of combinatorial shortcuts, we design a simple experiment on text classification tasks. As Figure~\ref{fig:demonstration} shows, we train attention models, where the query of attention is encoded with the whole sentences. However, we only allow attention to highlight keywords among the first five tokens and four default tokens, \ie, ``\texttt{A}'', ``\texttt{B}'', ``\texttt{C}'', and ``\texttt{D}''. After the \emph{Attention} module, the weighted sequences are further encoded to a fixed-length vector with the \emph{Encoder} module, and finally to predict the labels.
Since the default tokens appear in all samples and do not carry any meaningful information, if the attention mechanism can highlight the inputs' critical parts, little attention should be paid to them.
Moreover, we could check whether the models put differential attention to default tokens for different classes to show if the information is encoded in the mask due to combinatorial shortcuts.


We use the real-world IMDB dataset~\citep{maas2011learning} for experiments and examine different settings regarding the \emph{Encoder} module in Figure~\ref{fig:demonstration}, \ie, whether the encoder is a simple sum pooling or a position-aware trainable neural network model, \ie, recurrent convolutional neural network~(RCNN) proposed by \citet{lai2015recurrent}. 
We use pre-trained GloVe word embeddings~\citep{pennington2014glove} and keep them fixed to prevent shortcuts through word embeddings. We train soft attention models for 25 epochs with RMSprop optimizers using default parameters and record the averaged results of 10 runs. The results are reported in Table~\ref{tab:demonstration}. 

As we can see in Table~\ref{tab:demonstration}, the attention models placed more than half of the attention weights to the default tokens, and the weights for different classes were significantly different. 
Taking Pooling encoders as an example, the models put up to 68.0\% attention weights to default token ``\texttt{A}'' for positive samples, and put 36.6\%, 38.3\%, and 18.5\% attention weights to default token ``\texttt{B}'', ``\texttt{C}'', and ``\texttt{D}'' respectively for negative samples, summing up to 93.4\% in total. As for RCNN encoders, the results were similar.
The reason for observing repeatable results on default tokens with different initialization may be due to their slight asymmetry in the GloVe embedding space, \ie, ``\texttt{A}'' is slightly closer to ``\texttt{good}'' than to ``\texttt{bad}'', while the other three default tokens are just the opposite.
These results suggest that the attention mechanism may not work as expected to highlight the critical parts of inputs and provide interpretability. 
Instead, they learn to work as another kind of ``encoding layers'' and utilize the default tokens to fit the data through combinatorial shortcuts.

\section{Methods for Mitigating Combinatorial Shortcuts}
\label{sec:method}

In this section, based on the perspective of causal effect estimations introduced in Section~\ref{sec:gap_theory}, we come up with two methods, \emph{random attention pretraining} and \emph{mask-neutral learning with instance weighting} respectively, to mitigate combinatorial shortcuts for better interpretability. Then we compare the pros and cons of the two methods.

\subsection{Random attention pretraining}
\label{sec:pretrain}

We first propose a simple and straightforward method to address the issue. As analyzed in Section~\ref{sec:gap_theory}, the fundamental reason for combinatorial shortcuts is the biased estimation of $\mathbb{E}(Y|X \odot M)$, and random combinations of $X$ and $M$ can give unbiased results in theory. 

Inspired by this idea, we can generate the masks completely at random and train the attention model's downstream part. We then fix the downstream part, replace the random attention with a trainable attention layer, and train the attention layer only. As the downstream parts of neural networks are trained unbiasedly and fixed, solely training the attention layers is solving $m^*=\argmin_m \mathcal{L}\big(\mathbb{E}(Y|x \odot m), y\big)$ with an unbiased estimation of $\mathbb{E}(Y|X \odot M)$. Thus the interpretability is guaranteed.

\subsection{Mask-neutral learning with instance weighting}
The second method is based on instance weighting, which has been successfully applied for mitigating sample selection bias~\citep{zadrozny2004learning,zhang2019selection}, social prejudices bias~\citep{zhang2020demographics}, and also for recovering the causal effects~\citep{ertefaie2010comparing,winship1999estimation}. 
The core idea of this method is that instead of fitting a biased $\mathbb{E}(Y|X \odot M)$, with instance weighting, we could recover a \emph{mask-neutral distribution} where the masks are unrelated to the labels. Thus the downstream parts of attention layers become less biased, and combinatorial shortcuts can be partially mitigated.

\vspace{\paragraphskip} \noindent \textbf{Assumptions about the biased distribution and the mask-neutral distribution} \ 
We first define the mask-neutral distribution and its relationship with the biased distribution with which ordinary attention mechanisms are trained. 
Considering the downstream part of the attention layers, which estimates $\mathbb{E}(Y|X \odot M)$, we assume that there is a mask-neutral distribution $\mathscr{Q}$ with domain $\mathcal{X} \times \mathcal{Y} \times \mathcal{M} \times \mathcal{S}$, where $\mathcal{X}$ is the feature space, $\mathcal{Y}$ is the (binary) label space\footnote{We focus on binary classification problems in this paper. However, the proposed methodology can be easily extended to multi-class classifications.}, $\mathcal{M}$ is the feature mask space, and $\mathcal{S}$ is the binary sampling indicator space.
During the training procedure, the selective combination of masks and features result in combinatorial shortcuts. We assume for any given sample $(x, y, m, s)$ drawn independently from $\mathscr{Q}$, it will be selected to appear in the training of attention mechanisms if and only if $s=1$, which results in the biased distribution $\mathscr{P}$.
We use $P(\cdot)$ to represent probabilities of the biased distribution $\mathscr{P}$, and $Q(\cdot)$ for the mask-neutral distribution $\mathscr{Q}$, then we have
\begin{equation}
\label{eq:assumption1}
  P(\cdot) = Q(\cdot|S=1) \,.
\end{equation}

Ideally, we should have $M \perp (X,Y)$ on $\mathscr{Q}$ to obtain unbiased $\mathbb{E}(Y|X \odot M)$ as discussed in Section~\ref{sec:gap_theory}. However, when both sides are vectors, it will be intractable. Therefore, we take a step back and only assume $Y \perp M$ on $\mathscr{Q}$, \ie,
\begin{equation}
\label{eq:assumption2}
Q(Y|M)=Q(Y) \,.\footnote{We would like to note that this assumption does not mean that the ideal explanation $m^*$ is unrelated with $y$. This assumption only asks for the independent sampling of $M $ and $Y $ be simulated in the process of training the downstream parts of the attention layers, \ie, to encourage a relatively unbiased estimation of $\mathbb{E}(Y|X \odot M)$.}
\end{equation}

If $S$ is completely at random, $\mathscr{P}$ will be consistent with $\mathscr{Q}$. However, the attention layers are highly selective, making that only some combinations of $X$ and $M$ are visible to the downstream model. We assume that $ M$ and $Y$ control $S$. And for any given $Y$ and $M$, the probability of selection is greater than $0$, defined as
\begin{equation}
\label{eq:assumption3}
Q(S=1|X,Y,M) = Q(S=1|Y,M) > 0 \,.
\end{equation}

To further simplify the problem, we assume that the selection does not change the marginal probability of $M$ and $Y$, \ie,
\begin{equation}
\label{eq:assumption4}
P(M)=Q(M) \,,
P(Y)=Q(Y) \,.
\end{equation}
In other words, we assume that although $S$ is dependent on the combination of $M$ and $Y$ in $\mathscr{Q}$, it is independent on either $M$ or $Y$ only, \ie, $Q(S|M) = Q(S)$ and $Q(S|Y) = Q(S)$.

\vspace{\paragraphskip} \noindent \textbf{The unbiased expectation of loss with instance weighting} \ 
We show that, by adding proper instance weights, we can obtain an unbiased estimation of the loss on the mask-neutral distribution $\mathscr{Q}$, with only the data from the biased distribution $\mathscr{P}$.

\begin{theorem}[Unbiased Loss Expectation]
  \label{thm:unbiased}
  For any function $f=f(x \odot m)$, and for any loss $\mathcal{L} = \mathcal{L}\big(f(x \odot m ), y\big)$, if we use $w=\frac{P(y)}{P(y|m)}$ as the instance weights, then
    \begin{displaymath}
    \resizebox{\columnwidth}{!}{$
      \mathbb{E}_{x,y,m \sim \mathscr{P}} \Big[w\mathcal{L}\big(f(x \odot m), y\big) \Big] = \mathbb{E}_{x,y,m \sim \mathscr{Q}} \Big[ \mathcal{L}\big(f(x \odot m), y\big) \Big] \,.
    $}
    \end{displaymath}
\end{theorem}

Fact~\ref{thm:unbiased} shows that, by a proper instance-weighting method, the downstream part of the attention model can learn on the mask-neutral distribution $\mathscr{Q}$, where $Q(Y|M)=Q(Y)$.
Therefore, the independence between $M$ and $Y$ is encouraged, then it will be hard for the classifier to approximate $Y$ solely by $M$. Thus, the classifier will have to use useful information from $X$, and have the combinatorial shortcuts problem mitigated.

We present the proof for Fact~\ref{thm:unbiased} as follows.
\begin{proof}
  We first present an equation with the weight $w$,
    \begin{equation*}
      \begin{split}
        w=& \frac{P(y)}{P(y|m)} =  \frac{Q(y)}{Q(y|m, S=1)} \\
         =& \frac{Q(y)}{Q(S=1|y,m) Q(y|m) / Q(S=1|m)}\\
         =& \frac{Q(S=1)}{Q(S=1|y,m)} \\
         =& \frac{Q(S=1)}{Q(x,y,m|S=1)Q(S=1) / Q(x,y,m)} \\
         =& \frac{Q(x,y,m)}{P(x,y,m)}\,\text{.}
      \end{split}
    \end{equation*}
  Then we have
    \begin{displaymath}
      \begin{split}
          & \mathbb{E}_{x,y,m \sim \mathscr{P}} \Big[w\mathcal{L}\big(f(x \odot m), y\big) \Big] \\
        = & \int \frac{Q(x,y,m)}{P(x,y,m)} \mathcal{L}\big(f(x \odot m), y\big) dP(x,y,m) \\
        = & \int \mathcal{L}\big(f(x \odot m), y\big) dQ(x,y,m) \\
        = & \mathbb{E}_{x,y,m \sim \mathscr{Q}} \Big[ \mathcal{L}\big(f(x \odot m), y\big) \Big] \,.
      \end{split}
    \end{displaymath}
\end{proof}

\vspace{\paragraphskip} \noindent \textbf{Mask-neutral learning} \ 
With Fact~\ref{thm:unbiased}, we now propose mask-neutral learning for better interpretability of attention mechanisms. As shown, by adding instance weight $w=\frac{P(y)}{P(y|m)}$ to the loss function, we can obtain unbiased loss of the mask-neutral distribution. As distribution $\mathscr{P}$ is directly observable, estimating $P(\cdot)$ is possible. In practice, we could train a classifier to estimate $P(Y|M)$ along with the training of the attention layer, optimize it and the attention layers, as well as the other parts of models alternatively.

\begin{table*}[ht]
  \centering
  \caption{Effectiveness of the proposed methods for mitigating the combinatorial shortcuts.}
  \label{tab:mitigate}
  \vspace{\captionskip}
  \begin{tabular}{cccx{1.0cm}x{1.4cm}x{1.4cm}x{1.4cm}x{1.4cm}x{1.4cm}}
    \toprule
    \multirow{2}{*}{No.} & \multirow{2}{*}{Method} & \multirow{2}{*}{Encoder} & \multirow{2}{*}{Label} &    \multicolumn{5}{c}{Attention to default tokens} \\
    & & & & \texttt{A} & \texttt{B} & \texttt{C} & \texttt{D} & Total\\
    \midrule
    (1) & \multirow{4}{*}{Pretraining} & \multirow{2}{*}{Pooling} & \texttt{pos} & 0.0\% & 4.6\% & 2.3\% & 0.0\% & 6.9\% \\
    (2) & & & \texttt{neg} & 0.0\% & 0.4\% & 20.2\% & 0.0\% & 20.6\% \\
    (3) & & \multirow{2}{*}{RCNN} & \texttt{pos} & 0.2\% & 1.1\% & 1.0\% & 0.3\% & 2.6\% \\
    (4) & & & \texttt{neg} & 2.3\% & 4.3\% & 6.3\% & 2.8\% & 15.7\% \\
    \midrule
    (5) & \multirow{4}{*}{Weighting} & \multirow{2}{*}{Pooling} & \texttt{pos} & 1.3\% & 2.5\% & 0.5\% & 2.2\% & 6.5\% \\
    (6) & & & \texttt{neg} & 3.3\% & 0.7\% & 1.5\% & 1.2\% & 6.7\% \\
    (7) & & \multirow{2}{*}{RCNN} & \texttt{pos} & 6.7\% & 1.4\% & 10.3\% & 2.1\% & 20.5\% \\
    (8) & & & \texttt{neg} & 4.6\% & 1.8\% & 12.5\% & 1.9\% & 20.8\% \\
    \bottomrule
  \end{tabular}
\end{table*}

\subsection{Comparison of the two methods}
\label{sec:comparision}

As analyzed in Section~\ref{sec:gap_theory}, random attention pretraining is complete in theory. However, it may be practically incompetent because there are countless viable cases of the combinations of $X$ and $M$. It could be challenging to estimate $\mathbb{E}(Y|X \odot M)$ well, especially when the dimension of input features is high or strong co-adapting patterns exist in the data.
Under such cases, the pretraining procedure may become less efficient as it needs to explore all possible masks evenly, even if most of the masks are worthless. In conclusion, the model may fail to estimate complex $\mathbb{E}(Y|X \odot M)$ functions well in some cases, which finally limits the interpretability. 

Compared with the random attention pretraining method, the instance weighting-based approach concentrates more on the useful masks to address the shortcomings. Thus it will suffer less from the efficiency problem. Nevertheless, the effectiveness of the instance weighting method relies on the assumptions as shown in Equation~(\ref{eq:assumption1})--(\ref{eq:assumption4}). However, in some cases, the assumptions may not hold. For example, in Equation~(\ref{eq:assumption3}), we assume that given $Y$ and $M$, $S$ is independent on $X$. In other words, $X$ controls $S$ only through $Y$. This assumption is necessary for simplifying the problem, but may not sometimes hold when given $Y$ and $M$, $X$ can still influence $S$.
Besides, the effectiveness of the method also relies on an accurate estimation of $P(Y|M)$, which may require careful tuning as the probability $P(Y|M)$ is dynamically changing along the training process of attention mechanisms.

\section{Experiments}
\label{sec:experiments}

In this section, we present the experimental results of the proposed methods. For simplicity, we denote \emph{random attention pretraining} as \textbf{Pretraining} and \emph{mask-neutral learning with instance weighting} as \textbf{Weighting}. Firstly, we analyze the effectiveness of mitigating combinatorial shortcuts. Then, we examine the effectiveness of improving interpretability.

\subsection{Experiments for mitigating combinatorial shortcuts}
\label{sec:exp_shortcuts}

We applied the proposed methods to the experiments introduced in Section~\ref{sec:demo_design} to check whether we can mitigate the combinatorial shortcuts. We summarize the results in Table~\ref{tab:mitigate}.

As presented, after applying Pretraining and Weighting, the percentage of attention weights assigned to the default tokens were significantly reduced. Since that the default tokens do not provide useful information but only serve as carriers for combinatorial shortcuts, the results reveal that our methods have mitigated the combinatorial shortcuts successfully.

\subsection{Experiments for improving interpretability}
\label{sec:exp_interpretability}

\begin{table*}[ht]
  \centering
  \caption{Effectiveness of the proposed methods for improving interpretability. We report the post-hoc accuracy scores with different methods.}
  \label{tab:improve}
  \vspace{\captionskip}
  \begin{tabular}{cccx{1.8cm}x{1.8cm}x{1.8cm}x{1.8cm}}
    \toprule
    No. & \multicolumn{2}{c}{Method} & IMDB & Yelp~P. & MNIST & F-MNIST\\
    \midrule
    (1) & \multicolumn{2}{c}{Gradient~\citep{simonyan2013deep}} & 85.6\% & 82.3\% & 98.2\% & 58.6\% \\
    (2) & \multicolumn{2}{c}{LIME~\citep{ribeiro2016should}} & 89.8\% & 87.4\% & 80.4\% & 75.6\%\\
    (3) & \multicolumn{2}{c}{CXPlain~\citep{schwab2019cxplain}} & 90.6\% & 97.7\% & 99.4\% & 59.7\%
    \\
    \midrule
    (4) & \multicolumn{2}{c}{L2X~\citep{chen2018learning}} & 89.2\% & 88.2\% & 91.4\% & 77.3\% \\
    (5) & \multicolumn{2}{c}{VIBI~\citep{bang2019explaining}} & 90.8\% & 94.4\% & 98.3\% & 84.1\% \\
    (6) & \multicolumn{2}{c}{AIL~\citep{liang2020adversarial}$^\dagger$} & \textbf{98.5\%} & \textbf{99.3\%} & 99.0\% & \textbf{97.8\%} \\
    \midrule
    (7) & \multirow{3}{*}{L2X with $\hat{y}$}  & -- -- & 48.8\% & 77.8\% & 94.9\% & 85.3\% \\
    (8) & & Pretraining & {\bf 97.1\%} & {\bf 99.0\%} & 66.3\% & 89.4\% \\
    (9) & & Weighting & 94.3\% & 87.7\% & {\bf 99.8\%} & {\bf 95.4\%} \\
    \bottomrule
    \multicolumn{7}{r}{\small $^\dagger$AIL utilizes additional information about the models to be explained, \ie, their gradients.}\\
  \end{tabular}
\end{table*}

In this section, using L2X~\citep{chen2018learning} as an example and basis, we present the effectiveness of mitigating combinatorial shortcuts for better interpretability.

\subsubsection{Task and evaluation}

The task is the same as \citet{chen2018learning} and \citet{liang2020adversarial}, \ie, to find a small subset of input components that suffices on its own to yield the same outcome by the model to be explained, and the relative importance of each feature is allowed to vary across instances. We assume that we have access to the to-be-explained black-box model, and the task asks for the importance score of each feature component, and then the top $k$ most important feature components could be obtained. 
We use the same evaluation method as \citet{chen2018learning} and \citet{liang2020adversarial}, \ie, a predictive evaluation that assesses how accurate the to-be-explained model can approximate the original model-outputs using the selected feature components only, and we report this post-hoc accuracy. More details about the calculation of post-hoc accuracy are in Appendix~\ref{app:metrics}.

\subsubsection{Experimental settings}
Here we present the experimental settings. Due to space constraints, we present more details in Appendix~\ref{app:detailed_evaluation}.

\vspace{\paragraphskip} \noindent \textbf{Basic settings} \ 
Similar to~\citet{liang2020adversarial}, to further enrich the information for model explanations, we incorporate the to-be-explained model's outputs, \ie, $\hat{y}$, as part of the query for feature selection. 
As obtaining the outputs requires no further information apart from samples' features and the to-be-explained model, it does not hurt the model-agnostic property of explanation methods nor requires additional annotations.
We adopt binary feature-attribution masks to select features, \ie, top~$k$ values of the mask are set to $1$, others are set to $0$, then we treat $X \odot M$ as the selected features~\citep{chen2018learning}.
We repeated ten times with different initialization for each method on each dataset and report the averaged post-hoc accuracy results. 


\vspace{\paragraphskip} \noindent \textbf{Datasets} \ 
We report evaluations on four datasets: IMDB~\citep{maas2011learning}, Yelp~P.~\citep{zhang2015character}, MNIST~\citep{lecun1998gradient}, and Fashion-MNIST~(F-MNIST)~\citep{xiao2017fashion}.
IMDB and Yelp~P. are two text classification datasets. IMDB is with 25,000 train examples and 25,000 test examples. Yelp~P. contains 560,000 train examples and 38,000 test examples.
MNIST and F-MNIST are two image classification datasets. For MNIST, following~\citet{chen2018learning}, we collected a binary classification subset by choosing images of digits 3 and 8, with 11,982 train examples and 1,984 test examples. For F-MNIST, following \citet{liang2020adversarial}, we selected the data of Pullover and Shirt with 12,000 train examples and 2,000 test examples.

\vspace{\paragraphskip} \noindent \textbf{Models to be explained} \ 
The same as \citet{chen2018learning}, for IMDB and Yelp~P., we implemented CNN-based models and selected 10 and 5 words, respectively, for explanations. For MNIST and F-MNIST, we used the same CNN model as \citep{chen2018learning} and selected 25 and 64 pixels, respectively~\citep{liang2020adversarial}.

\vspace{\paragraphskip} \noindent \textbf{Baselines} \ 
We considered state-of-the-art model-agnostic baselines:
LIME~\citep{ribeiro2016should}, CXPlain~\citep{schwab2019cxplain}, L2X~\citep{chen2018learning}, VIBI~\citep{bang2019explaining}, and AIL~\citep{liang2020adversarial}.
We also compared with model-specific baselines, \ie, Gradient~\citep{simonyan2013deep}. Our methods follow the same paradigm as L2X, VIBI, and AIL. A brief introduction to the baseline methods can be found in Appendix~\ref{app:baselines}.

\begin{table*}[ht]
  \centering
   \caption{The example results of explanation from IMDB. The model to be explained makes the incorrect prediction for this sample, \ie, the ground-truth label of the review text is \texttt{negative}, while the trained model predicts \texttt{positive} for it.}
  \label{tab:imdb_example}
  \vspace{\captionskip}
  \begin{tabular}{x{2cm}|p{14cm}}
    \toprule
    L2X & \scriptsize Aussie Shakespeare for 18-24 set. With blood, blood and more blood, and good dose of nudity. This will not be for every one on \hl{may} levels, to violent for some too cheap for most. Done on low budget they try and do there best but it \hl{only} works sporadically. And this macbeth just \hl{seem} to be lacking, it's just \hl{not} compelling. Although there is some good acting on the part of most you don't get into there heads especially mecbeths. The \hl{best} performance came from Gary sweet and the strangest mick molly. If your into Shakespeare then see it, but if you like your cheese mature you will love it. It not a bad film but it not that good either. Sam Peckenpah would of loved it, that is if it was filmed as a western. I was expecting a lot from this, as I loved romper stomper. But this \hl{is was a vacant effort}. \quad (Prediction with key words: \texttt{negative})\\
    \midrule
    L2X~with~$\hat{y}$ & \scriptsize \hl{Aussie} Shakespeare for 18-24 set. With blood, blood \hl{and} more blood, \hl{and} good \hl{dose} of nudity. This will not be for every one on may levels, to violent for some too cheap for most. Done on low budget they try \hl{and} do there best but it only works \hl{sporadically}. And this \hl{macbeth} just seem to be lacking, it's just not compelling. Although there is some good acting on the part of most you don't get into there heads especially mecbeths. The best performance came from Gary sweet \hl{and} the strangest mick molly. If your into Shakespeare then see it, but if you like your cheese mature you will love it. It not a bad film but it not that good either. Sam Peckenpah would of loved it, that is if it was filmed as a western. I was expecting a lot from this, as I loved \hl{romper} stomper. But this is was a \hl{vacant} effort. \quad (Prediction with key words: \texttt{positive})\\
    \midrule
    L2X~with~$\hat{y}$ (Pretrain) & \scriptsize Aussie Shakespeare for 18-24 set. With blood, blood and more blood, and good dose of nudity. This will not be for every one on may levels, to violent for some too cheap for most. Done on low budget they try and do there \hl{best} but it only works sporadically. And this macbeth just seem to be lacking, it's just not compelling. Although there is some good acting on the part of most you don't get into there heads \hl{especially} mecbeths. The \hl{best performance came} from Gary sweet and the strangest mick molly. If your into Shakespeare then see it, but if you like your cheese mature you will \hl{love} it. It not a bad film but it not that good either. Sam Peckenpah would of \hl{loved} it, \hl{that} is if it was filmed as a western. I was expecting a lot from this, \hl{as} I \hl{loved} romper stomper. But this is was a vacant effort. \quad (Prediction with key words: \texttt{positive})\\
    \midrule
    L2X~with~$\hat{y}$ (Weight) & \scriptsize Aussie Shakespeare for 18-24 set. With blood, blood and more blood, and good dose of nudity. This will not be for every one on \hl{may} levels, to violent \hl{for} some too cheap for most. Done on low budget they \hl{try} and do there best but it only works sporadically. And this macbeth just \hl{seem} to be lacking, \hl{it's} just not compelling. Although there is some \hl{good} acting on the part of most \hl{you} don't get into there heads especially mecbeths. The best performance came from Gary sweet and the strangest mick molly. If your into Shakespeare then see it, but if \hl{you like} your cheese mature you will love it. It not a bad film but it not that good either. Sam Peckenpah would of loved it, that is if it was filmed as a western. I was expecting a lot from this, as I loved romper stomper. But \hl{this} is was a vacant effort. \quad (Prediction with key words: \texttt{positive})\\
    \bottomrule
  \end{tabular}
\end{table*}

\begin{table*}[ht]
  \centering
  \caption{The example results of explanation from Yelp P. The model to be explained makes the correct prediction for this sample, \ie, the ground-truth label of the review text is \texttt{positive}, and the trained model predicts \texttt{positive} for it.}
  \label{tab:yelp_example}
  \vspace{\captionskip}
  \begin{tabular}{x{2cm}|p{14cm}}
    \toprule
    L2X & \scriptsize Good choice if \hl{you} are looking for \hl{a} pricier Italian menu. They feature veal entrees and a pretty good assortment of Italian seafood dishes. Their homemade gnocchi and cavatelli are wonderful as well. The service is \hl{typically} so-so, but the great food keeps me coming back. \quad (Prediction with key words: \texttt{positive})\\
    \midrule
    L2X~with~$\hat{y}$ & \scriptsize Good choice if you are looking for a pricier Italian menu. They feature veal entrees and a pretty good assortment of Italian seafood dishes. Their homemade gnocchi and cavatelli are wonderful \hl{as well}. The service is typically \hl{so-so}, but the great food keeps me coming back. \quad (Prediction with key words: \texttt{positive})\\
    \midrule
    L2X~with~$\hat{y}$ (Pretrain) & \scriptsize Good choice if you are looking for a pricier Italian menu. They feature veal entrees and a pretty good assortment of Italian seafood \hl{dishes}. Their \hl{homemade} gnocchi and cavatelli are \hl{wonderful} as \hl{well}. The service is typically so-so, but the \hl{great} food keeps me coming back. \quad (Prediction with key words: \texttt{positive})\\
    \midrule
    L2X~with~$\hat{y}$ (Weight) & \scriptsize \hl{Good choice} if you are \hl{looking} for a pricier Italian menu. They feature veal entrees and a pretty good assortment of Italian seafood dishes. Their \hl{homemade} gnocchi and cavatelli are wonderful as well. The service is typically so-so, but the \hl{great} food keeps me coming back. \quad (Prediction with key words: \texttt{positive})\\
    \bottomrule
  \end{tabular}
\end{table*}

\subsubsection{Experimental results}
\label{sec:experimental_results}

Following the aforementioned evaluation scheme, we report the results in Table~\ref{tab:improve}.

From the table, we can find that directly adding $\hat{y}$ to the query did not always improve the performance by comparing Row~(4) and (7). Interestingly, for the text classification datasets, adding $\hat{y}$ led to decreased performance, and meanwhile, Pretraining outperformed Weighting. For the image classification datasets, we had the exact opposite conclusion.
We ascribe this phenomenon to the inherent differences between the two tasks. Intuitively, people could easily guess the emotional polarity of a sentence from a few randomly selected words in a sentence, but it is not easy to guess the content of a picture from a few randomly selected pixels in an image. The function of $\mathbb{E}(Y|X \odot M)$ may be smoother and more comfortable to learn for text classification tasks than for image classification tasks.
As a result, as discussed in Section~\ref{sec:comparision}, it may be hard for Pretraining to learn reasonable estimations of $\mathbb{E}(Y|X \odot M)$ efficiently for images. Thus the performance of interpretability is limited, especially for MNIST, where the digital numbers are placed randomly, compared with F-MNIST, where the items are aligned better.

By comparing with the baselines~(especially L2X with $\hat{y}$), we find that despite the simplicity, Pretraining and Weighting can outperform most of the baselines and give comparable results with AIL, which is state of the art and utilizes additional information of the to-be-explained models, \ie, their gradients. We conclude that mitigating the combinatorial shortcuts can effectively improve interpretability.

\subsubsection{Visualization examples of explanation}
Here, we present the interpretation results of the examples from the tested datasets. Please note that we are trying to explain how machine learning models make decisions. These models are trained on datasets of limited size, and the datasets may be more or less biased, so the interpretation results may be inconsistent with ordinary people's understanding, especially for images.

\vspace{\paragraphskip} \noindent \textbf{IMDB} \ 
We first present an example from IMDB dataset. The results are in Table~\ref{tab:imdb_example}. Explanations to the functional tokens like \texttt{<START>} and \texttt{<PAD>} are omitted.

As shown from the example, L2X with $\hat{y}$ tends to select high-frequency but meaningless words~(\eg, \emph{and}), or niche words which are merged as \texttt{<UNK>} token~(\eg, \emph{Aussie} and \emph{sporadically}). This phenomenon explains the reason why L2X with $\hat{y}$ get poor post-hoc accuracy scores during our experiments on text classification datasets, and as analyzed in Section~\ref{sec:problem}, this can be ascribed to the combinatorial shortcuts. After applying the proposed methods~(\ie, Pretrain and Weight), the combinatorial shortcut problem has been mitigated successfully, and the explaining model can select better words for explanations. Besides, we can find that after adding the information of $\hat{y}$ to the query, the explanation models could better capture the key parts corresponding to the original given model to be explained. For example, L2X selects ``\emph{is was a vacant effort}'' in Table~\ref{tab:imdb_example} and the given model predicts \texttt{negative} with the selected words by L2X, however, the model to be explained actually predicts \texttt{positive} for this example, and after adding $\hat{y}$ and applying the proposed methods, the explanation results become more consistent to the given model.

\vspace{\paragraphskip} \noindent \textbf{Yelp P.} \ 
In this section, we present an example from Yelp P. dataset. The results are in Table~\ref{tab:yelp_example}. Similarly, explanations to the functional tokens like \texttt{<START>} and \texttt{<PAD>} are omitted.

Compared with the results of IMDB, we find that the explanation results of L2X seem to be more susceptible to the combinatorial shortcut problem, as words like \emph{a} and tokens like \texttt{<PAD>} appear in the explanation results of L2X more frequently. This phenomenon may be ascribed to that the sentences in Yelp P. are shorter than IMDB, thus the explainer could more easily capture the global meaning of the sentences and form combinatorial shortcuts. For L2X with $\hat{y}$, the results are consistent with the results of IMDB, \ie, meaningless words and \texttt{<UNK>} tokens are frequently selected. Pretrain and Weight can help mitigate the problem of combinatorial shortcuts, and thus improve the performance of interpretability.

\begin{figure}
    \centering
    \includegraphics[width=0.95\columnwidth]{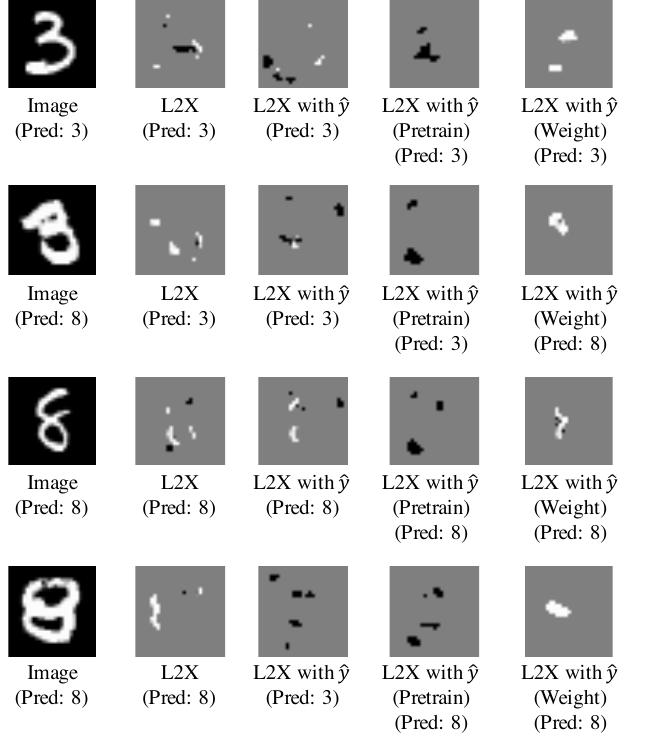}
    \vspace{-2em}
    \caption{The example result of explanation on images from MNIST. The 4th row is the least confident sample of number ``8''.}
    \label{fig:mnist_example}
\end{figure}

\vspace{\paragraphskip} \noindent \textbf{MNIST} \ 
Figure~\ref{fig:mnist_example} shows four examples from MNIST. We find that the given model to be explained predicts all samples with label \texttt{8} in the test set correctly, so we choose the least confident sample to present in the 4th row of Figure~\ref{fig:mnist_example}.

The gray parts of images are unselected pixels, while the white/black parts are selected pixels. We find a very significant tendency that L2X and Weight tend to select white pixels in the image, while L2X with $\hat{y}$ and Pretrain tend to select black pixels. In the 2nd row of Figure~\ref{fig:mnist_example}, only Weight select the right parts which makes the model to be explained predict \texttt{8} for this sample.

\begin{figure}
    \centering
    \includegraphics[width=0.95\columnwidth]{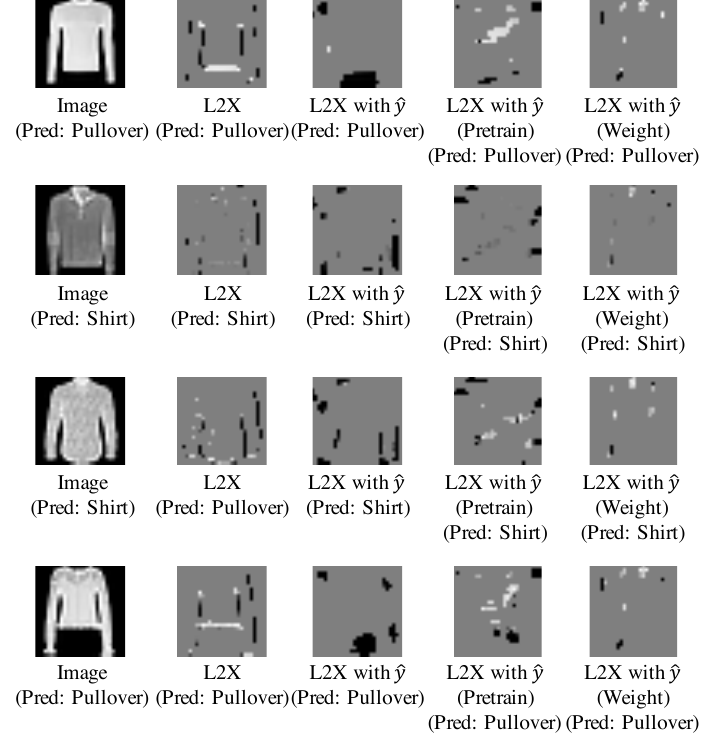}
    \vspace{-2em}
    \caption{The example result of explanation on images from Fashion-MNIST.}
    \label{fig:fmnist_example}
\end{figure}

\vspace{\paragraphskip} \noindent \textbf{F-MNIST} \ 
Figure~\ref{fig:fmnist_example} shows four examples from Fashion-MNIST. Similarly, the gray parts of images are unselected pixels, and the white/black parts are selected pixels.

Interestingly, we find that L2X tends to select the pixels on the edges. Similar to the examples of L2X in Table~\ref{tab:imdb_example}, the highlighted parts given by L2X may tend to be important generally, however, it may not be the most significant part for explaining the predictions of given models. On the other side, the explanations given by Weight tend to be of the neckline parts and the shoulder parts of the clothing, and maybe more informative to explain the predictions of machine learning models. For example, in the 3rd row of Figure~\ref{fig:fmnist_example}, L2X selects the edges of the given image, while the model to be explained judges that the select features suggest that the image belongs to a \texttt{Pullover}.

\section{Conclusion}
\label{sec:conclusion}

Attention-based model interpretations have been popular for their convenience to integrate with neural networks. However, many researchers find that attention sometimes yields non-interpretable results, and there has been a debate on the interpretability of the attention mechanisms. This paper proposes that the combinatorial shortcuts are one of the root causes hindering attention mechanisms' interpretability. We analyze the combinatorial shortcuts theoretically and design experiments to show their existence. Furthermore, we propose two methods to mitigate combinatorial shortcuts for better interpretability. Experiments show that the proposed methods effectively mitigate the adverse impacts and improve the interpretability of attention mechanisms. The results presented in this paper can help us better understand how attention mechanisms work.

\bibliographystyle{ACM-Reference-Format}
\bibliography{references}

\clearpage

\appendix

\section{Details about the Evaluation Scheme}
\label{app:detailed_evaluation}

Here we present the structure of the models used for our experiments, as well as how the evaluation metrics are calculated.

\subsection{The text classification models for IMDB and Yelp~P.}

\begin{table}[ht]
  \centering
  \caption{The structure of text classification models for IMDB and Yelp~P.}
  \label{tab:text_model}
  \resizebox{\columnwidth}{!}{
  \begin{tabular}{cccccc}
    \toprule
    Layer & \# Units & Kernel Size & Stride & Padding & Activation\\
    \midrule
    Embedding & 50 & -- -- & -- -- & -- -- & -- -- \\
    Dropout~($0.2$) & -- -- & -- -- & -- -- & -- -- & -- -- \\
    Convolution~(1D) & 250 & 3 & 1 & default & ReLU \\
    GlobalMaxPooling & -- -- & -- -- & -- -- & -- -- & -- -- \\
    Fully-Connected & 250 & -- -- & -- -- & -- -- & ReLU \\
    Dropout~($0.2$) & -- -- & -- -- & -- -- & -- -- & -- -- \\
    Fully-Connected & 2 & -- -- & -- -- & -- -- & Softmax \\
    \bottomrule
  \end{tabular}
  }
\end{table}

The structure of neural network models to be explained for text classification datasets~(including IMDB and Yelp~P.) is presented in Table~\ref{tab:text_model}. This model structure achieved accuracy scores of about 96.1\% and 89.0\% on the train set and the test set of IMDB respectively, and about 96.0\% and 95.4\% on the train set and test set of Yelp~P. respectively. The structures of explainers, as well as the approximators of models including L2X, VIBI, AIL, and L2X with $\hat{y}$ are slightly different from this to incorporate global information when selecting features and to select keywords. During our experiments, we find that while Pretrain prefers powerful approximators, Weight works more stable with simple approximators. This result is consistent with the results in Table~\ref{tab:mitigate}. Besides, as estimating $\mathbb{E}(Y|X \odot M)$ requires good calibrations, layers that act differently during training and testing, like \emph{Dropout} and \emph{BatchNormalization} layers, may not be welcomed by explanation models.

\subsection{The image classification model for MNIST and Fasion-MNIST}

\begin{table}[ht]
  \centering
  \caption{The structure of image classification models for MNIST and Fashion-MNIST}
  \label{tab:image_model}
  \resizebox{\columnwidth}{!}{
  \begin{tabular}{cccccc}
    \toprule
    Layer & \# Units & Kernel Size & Stride & Padding & Activation\\
    \midrule
    Convolution~(2D) & 32 & (3, 3) & 1 & default & ReLU \\
    Convolution~(2D) & 64 & (3, 3) & 1 & default & ReLU \\
    MaxPooling~(2D) & -- -- & (2, 2) & (2, 2) & -- -- & -- -- \\
    Flatten & -- -- & -- -- & -- -- & -- -- & -- -- \\
    Dropout~($0.25$) & -- -- & -- -- & -- -- & -- -- & -- -- \\
    Fully-Connected & 128 & -- -- & -- -- & -- -- & ReLU \\
    Dropout~($0.5$) & -- -- & -- -- & -- -- & -- -- & -- -- \\
    Fully-Connected & 2 & -- -- & -- -- & -- -- & Softmax \\
    \bottomrule
  \end{tabular}
  }
\end{table}

The structure of neural network models to be explained for image classification datasets~(including MNIST and F-MNIST) is presented in Table~\ref{tab:image_model}. This model structure achieved accuracy scores of more than 99.8\% on both the train set and test set of MNIST, and about 96.1\% and 92.2\% on the train set and test set of F-MNIST respectively. The structures of explainers, as well as the approximators of models including L2X, VIBI, AIL, and L2X with $\hat{y}$, are slightly different from this to incorporate global information and to select pixels.

\subsection{Calculation of the evaluation metrics}
\label{app:metrics}

The same with \citet{chen2018learning} and \citet{liang2020adversarial}, we used \emph{post-hoc accuracy} for quantitatively validating the effectiveness of the methods. After feature selection, unselected words of texts were filled with \texttt{<PAD>} tokens, and unselected pixels of images were filled with their average values. Then we used the model to be explained to predict the labels and compared whether the labels change or not. If the model makes consistent predictions before and after feature selection, the selected features may be informative for the model to make the decisions.

\subsection{Brief introduction to the baseline methods}
\label{app:baselines}
Among the baseline methods, Gradient~\citep{simonyan2013deep} takes advantage of the property of neural networks and selects the input features which have the most significant absolute values of gradients. 
LIME~\citep{ribeiro2016should} explains a model by quantifying the model's sensitivity to changes in the input. CXPlain~\citep{schwab2019cxplain} involves the real labels $y$ to compute the loss-function values by erasing each feature to zero and normalizes the loss-function values as the surrogate for ideal explanations for a neural network model to learn. Our methods follow the same paradigm as L2X~\citep{chen2018learning}, VIBI~\citep{bang2019explaining}, and AIL~\citep{liang2020adversarial}, which use hard attention to select a fixed number of features to approximate the output of the original models to be explained. VIBI improves L2X to encourage the briefness of the learned explanation by adding a constraint for the feature scores to a global prior. AIL~\citep{liang2020adversarial} is the state-of-the-art method that use adversarial training to encourage the gap between the predictability of selected/unselected features, additionally, AIL incorporate the outcome it aims to justify explicitly, as well as the gradients of the original model w.r.t. the samples to provide a warm start.

\subsection{Comparison with adversarial solutions for improving interpretability}

\citet{liang2020adversarial} and \citet{yu2019rethinking} also mentioned the phenomenon that attention could find a degenerated solution where attention could highlight feature components at different positions for different classes, and then the downstream parts of models may predict the label through this pattern, resulting in the lack of interpretability. And they used adversarial training to indirectly mitigate this combinatorial shortcut problem. However, they did not analyze why combinatorial shortcuts exist, and the adversarial scheme may change the behavioral tendency of models. For example, considering a sample with features $\{a, a', b, b', \text{noisy features}\}$, where $a'$ is collinear with $a$, $b'$ is collinear with $b$, and $a$ is relatively more predictive than $b$ while they are somehow complementary, the adversarial methods that encourage the gap between the predictability of selected/unselected features may tend to select $a$ and $a'$ to maximum the gap, and fail to select $a$ and $b$ which might be more interpretable. 

\end{document}